%% file: main.tex
\pdfoutput=1 % arXiv: force pdfLaTeX
\documentclass[conference]{IEEEtran}
\IEEEoverridecommandlockouts

\usepackage[T1]{fontenc}
\usepackage[utf8]{inputenc}
\usepackage{cite}
\usepackage{amsmath,amssymb,amsfonts}
\usepackage{float}        % must come BEFORE algorithm to enable [H] placement
\usepackage{algorithmic}
\usepackage{algorithm}
\usepackage{placeins}     % \FloatBarrier
\usepackage{afterpage}    % deferred float placement
\usepackage{graphicx}
\usepackage{textcomp}
\usepackage[table]{xcolor}
\usepackage{booktabs}
\usepackage{multirow}
\usepackage{array}
\usepackage{makecell}
\usepackage{subcaption}
\usepackage{hyperref}
\usepackage{url}
\usepackage{microtype}

\usepackage{eso-pic}
\newcommand\copyrightnotice{%
  \AddToShipoutPictureBG*{\AtPageLowerLeft{\makebox[\paperwidth]{\raisebox{1.5em}{%
    \parbox{0.85\paperwidth}{\centering\scriptsize
    \copyright~2026 IEEE. Personal use of this material is permitted. Permission from IEEE must be obtained for all other uses, in any current or future media, including reprinting/republishing this material for advertising or promotional purposes, creating new collective works, for resale or redistribution to servers or lists, or reuse of any copyrighted component of this work in other works.}}}}}}
\hypersetup{
    colorlinks=true,
    linkcolor=blue!60!black,
    citecolor=blue!60!black,
    urlcolor=blue!60!black
}

\def\BibTeX{{\rm B\kern-.05em{\sc i\kern-.025em b}\kern-.08em
    T\kern-.1667em\lower.7ex\hbox{E}\kern-.125emX}}

\usepackage{xspace}
\input{00_macros}

\begin{document}

% ---------- Title ----------
\title{CASCADE: Constraint-Aware SCAffolded DEcomposition for Logical Anomaly Detection}

% ---------- Authors (camera-ready) ----------
% \author{
% \IEEEauthorblockN{\rev{Jinchan Kim}}
% \IEEEauthorblockA{\rev{\textit{Ajou University}}\\
% \rev{Suwon, Republic of Korea}\\
% \rev{jinchank0959@ajou.ac.kr}}
% \and
% \IEEEauthorblockN{\rev{Jinhee Im}}
% \IEEEauthorblockA{\rev{\textit{Ajou University}}\\
% \rev{Suwon, Republic of Korea}\\
% \rev{jinhee8553@ajou.ac.kr}}
% \and
% \IEEEauthorblockN{\rev{Hyunsouk Cho\textsuperscript{*}}}
% \IEEEauthorblockA{\rev{\textit{Ajou University}}\\
% \rev{Suwon, Republic of Korea}\\
% \rev{hyunsouk@ajou.ac.kr}}
% \thanks{\rev{\textsuperscript{*}Corresponding author.}}%
% \thanks{\revb{Our code and reproduction materials are available at \url{https://github.com/edjjincode/CASCADE}.}}%
% }

\author{
\IEEEauthorblockN{Jinchan Kim}
\IEEEauthorblockA{\textit{Dept. of Artificial Intelligence}\\
\textit{Ajou University}\\
Suwon, Republic of Korea\\
jinchank0959@ajou.ac.kr}
\and
\IEEEauthorblockN{Jinhee Im}
\IEEEauthorblockA{\textit{Dept. of Artificial Intelligence}\\
\textit{Ajou University}\\
Suwon, Republic of Korea\\
jinhee8553@ajou.ac.kr}
\and
\IEEEauthorblockN{Hyunsouk Cho\textsuperscript{*}}
\IEEEauthorblockA{\textit{Dept. of Artificial Intelligence}\\
\textit{Ajou University}\\
Suwon, Republic of Korea\\
hyunsouk@ajou.ac.kr}
\thanks{\textsuperscript{*}Corresponding author.}%
% \thanks{\revb{\url{https://github.com/edjjincode/CASCADE}.}}%
}

\maketitle
\copyrightnotice
% \linenumbers   % camera-ready: 줄번호 제거
% ---------- Abstract ----------
\begin{abstract}
Logical Anomaly Detection (LAD) requires verifying whether all category-specific constraints, such as object counts, pairing relations, and measurement ranges, are satisfied by a query image given only a few normal references.
Recent LAD methods based on Large Vision-Language Models (LVLMs) formulate this as checklist verification, decomposing the decision into independent textual checks. This decomposition, however, is text-only. Every item still reads the same unmodified query image, so the LVLM must extract different forms of visual evidence from one raw image. We identify this text--image mismatch as a constraint-specific perception gap, the source of two failure modes: instance conflation for binding constraints and scale drift for calibration constraints.
To diagnose whether these failures stem from insufficient reasoning or insufficient perception, we compare textual reasoning guides and visual perception guides in a training-free setting.
Our analysis shows that reasoning over the same image provides little benefit, whereas adapting the visual input to each constraint substantially improves verification accuracy.
Based on this finding, we propose \ours, a training-free framework that extends checklist decomposition using constraint-specific visual scaffolds.
We further introduce \ourspp\ which adapts these scaffolds at the object level by deriving their configurations from the constraint structure and reference distribution.
Across MVTec-LOCO and VID-AD, \ourspp\ surpasses the full-shot state of the art by $3.9$ AUROC points with only three references, and improves over the strongest VID-AD baseline by $11.4$, transferring across multiple LVLM backbones without any training.
Our code and reproduction materials are publicly  available\footnote{\url{https://github.com/edjjincode/CASCADE}}.
\end{abstract}


\input{table/main_loco}
\input{sections/01_introduction_new}
\input{sections/02_related_work}

\input{sections/03_preliminaries}
\input{sections/04_method_new}
\input{sections/05_analysis_new}
\input{sections/06_SOLVE}

\input{sections/07_SOLVE++}
\input{sections/08_new_experiment}

\input{sections/09_conclusion}

% Flush every experiment float (esp. wide tab:main-loco) before Conclusion,
% so no body table can drift past the appendix boundary.
\FloatBarrier

% \input{sections/07_conclusion}

% ---------- Acknowledgments (camera-ready) ----------
\section*{Acknowledgment}
This work was partly supported by Institute of Information \&
communications Technology Planning \& Evaluation (IITP) under the Artificial
Intelligence Convergence Innovation Human Resources Development
(IITP-2026-RS-2023-00255968) grant funded by the Korea government (MSIT), IITP
grant funded by the Korea government (MSIT) (No.~RS-2022-II220680, Abductive
inference framework using omni-data for understanding complex causal relations;
RS-2026-25617480), the Nano \& Material Technology Development Program through
the National Research Foundation of Korea (NRF) funded by Ministry of Science
and ICT (RS-2024-00444182), and the National Research Foundation of Korea (NRF)
grant funded by the Korea government (MSIT) (RS-2026-25518808).

% ==============================================================
\bibliographystyle{IEEEtran}
\bibliography{references}

% ==============================================================
% Appendix removed: the 10-page ICDM limit includes references and appendices.
% The multi-group solver is covered by one sentence in Sec. VII-B and by the released code.
\FloatBarrier

\end{document}

%% file: 00_macros.tex
\newcommand{\method}{CASCADE}
\newcommand{\ours}{\textbf{CASCADE}}
\newcommand{\ourspp}{\textbf{CASCADE++}}

\newif\ifshowrev
\showrevfalse
\newcommand{\rev}[1]{\ifshowrev\textcolor{blue}{#1}\else#1\fi}

\definecolor{revpurple}{rgb}{0.55,0.0,0.70}
\newif\ifshowrevb
\showrevbfalse
\newcommand{\revb}[1]{\ifshowrevb\textcolor{revpurple}{#1}\else#1\fi}
\providecommand{\revc}[1]{#1}

%% file: sections/01_introduction_new.tex
\section{Introduction}
\label{sec:intro}

Driven by recent advances in vision-language models, Logical Anomaly
Detection (LAD)~\cite{bergmann2022beyond} has emerged as an
important task in manufacturing inspection. Traditional anomaly
detection~\cite{roth2022patchcore, batzner2024efficientad}
focuses on identifying pixel-level defects such as cracks,
scratches, and contamination. LAD, by contrast, requires models
to verify logical constraints, such as object counts, pairing
relationships, or measurement ranges, from only a few normal
references and a query image. Since
violating even a single constraint renders the instance
anomalous, every constraint must be verified without exception.

\begin{figure}[!t]
\centering
\includegraphics[width=1\linewidth]{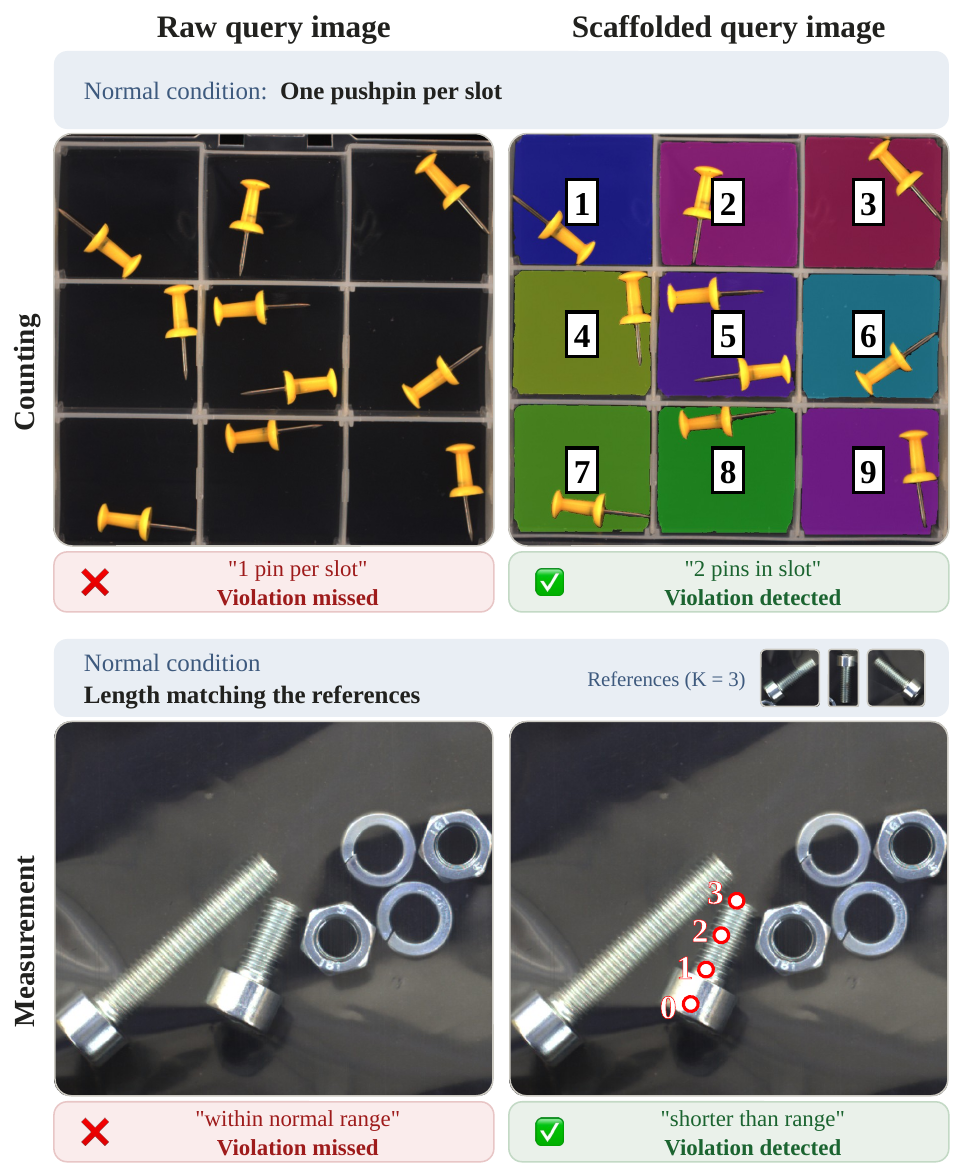}

\caption{
Checklist verification on a raw query versus a scaffolded query---per-instance coloring for counting, an object-anchored grid for measurement.
Both are the same image verified against the same checklist item, and both violate their constraint, yet only the scaffolded query yields the correct verdict.
The Measurement condition is defined by the $K\!=\!3$ references shown in the band rather than stated explicitly (Section~\ref{subsec:solvepp-grid}).
}
\label{fig:teaser}
\end{figure}

To verify the validity of a query image with only a few normal references, recent approaches have adopted Large Vision-Language Models (LVLMs) by formulating the task as a checklist evaluation.
These LVLM-based LAD methods~\cite{kwon2025logicqa,jin2025logicad} decompose the overall verification into independent per-constraint checks by explicitly converting constraints into a checklist.
First, in the generation phase, each logical constraint is converted into a textual checklist item that an LVLM can evaluate.
Then, in the verification phase, each checklist item is verified against the query image individually.
This checklist-based formulation is useful because it reduces the chance of overlooking a constraint violation and makes it possible to trace which constraint was violated.

This decomposition, however, creates a mismatch between text
and image. The checklist tailors the question to each constraint,
yet every check still reads the same unmodified raw query image.
The LVLM must therefore extract different types of visual evidence
from an identical input images, despite the
fact that different constraints demand fundamentally different
perceptual operations. Count and pairing constraints require the
model to individuate and bind visually similar instances, whereas
length, ratio, and fill-level constraints require the model to
compare continuous measurements against reference-defined normal
ranges. This mismatch manifests as two recurring failure modes in
LAD: instance conflation, where visually similar objects are not
reliably separated, and scale drift, where measurement judgments
become inconsistent across references and queries. To determine
whether these failures stem from insufficient reasoning or
insufficient perception, a direct diagnosis is needed.

To diagnose this under the training-free setting, we compare two
input-level interventions that require no model update. The first
is a textual reasoning guide~\cite{wei2022chain, madaan2023self}
that keeps the image unchanged but prompts more deliberate
reasoning over each checklist item. The second is a visual
perceptual guide~\cite{shtedritski2023does, lei2025scaffold,
izadi2026visual} that modifies the image itself by making
constraint-relevant visual information explicit. A textual
guide that resolves the errors would locate the limitation in
reasoning; a visual guide that does so would locate it in
perception.

Our analysis confirms that the limitation is perceptual, and that
the visual input must be adapted per constraint alongside the
checklist. As illustrated in Figure~\ref{fig:teaser}, a scaffolded query makes the
required instances and measurement bases explicit, enabling the model to
successfully detect the anomalies.

The scaffold must be constraint-specific, as
different constraints demand different forms of perception. Binding
constraints require separating and associating visually similar
instances, which mark-based scaffolds~\cite{yang2023setofmark} make
explicit, whereas calibration constraints require comparing
measurements against reference-defined normal ranges, for which
coordinate-based scaffolds~\cite{lei2025scaffold} supply a frame.

Based on this principle, we propose
\ours, a training-free framework that selects the scaffold type
from the constraint and overlays it on the query image, so that each
check reads a view adapted to what it examines. \ours\ alone,
however, leaves the scaffold's configuration at a fixed default, and
a default that misaligns with the constraint can mislead the
verifier more than no scaffold at all. We therefore extend it to
\ourspp, which derives each scaffold's configuration from the
constraint's logical structure and the reference distribution.
Our contributions are summarized as follows.
\begin{itemize}
\item We diagnose the limitation of checklist-based LVLM verification by comparing textual reasoning guides and visual perceptual guides, showing that LAD failures stem from insufficient constraint-relevant visual perception rather than from insufficient reasoning.
We propose \ours, a training-free framework that routes each constraint to its matched scaffold type, and extend it to \ourspp, which derives the scaffold's configuration from the constraint's logical structure and the reference distribution.

\item \revc{With three references and no training, \ourspp\ exceeds the full-shot state of the art by $3.9$ AUROC on MVTec-LOCO and the strongest VID-AD baseline by $11.4$, transferring across four LVLM backbones.}
\end{itemize}

%% file: sections/02_related_work.tex
\section{Related Work}
\label{sec:related}

\subsection{Vision-based LAD}
Vision-based methods detect anomalies as deviations from learned
normal patterns. Full-shot methods such as
PatchCore~\cite{roth2022patchcore} and
EfficientAD~\cite{batzner2024efficientad} achieve strong structural
anomaly detection but degrade on logical anomalies.
CSAD~\cite{hsieh2024csad} and SALAD~\cite{fucka2025salad} address this
gap by modeling component structure, yet they still require large
training sets per category.
Training-free methods such as
SAM-LAD~\cite{peng2025samlad} and UniVAD~\cite{gu2025univad}
alleviate this requirement by leveraging foundation models.
Despite these advances, all vision-based methods
share a fundamental limitation: they output only a scalar or
matching score, and cannot identify which constraint was violated or
explain why.

\subsection{LVLM-based LAD}
Recent methods shift from deviation measurement to direct constraint verification by feeding natural-language constraints to LVLMs.
LogicAD~\cite{jin2025logicad} and LogicQA~\cite{kwon2025logicqa} convert constraints into textual checklist items,
LogiCode~\cite{zhang2024logicode} compiles them into executable programs,
and LogSAD~\cite{zhang2025logsad} routes the task through multiple vision-language foundation models, together enabling traceable constraint-level verification.
Across these approaches, however, the decomposition operates on the text side alone.
The check is tailored to the constraint, but every check still reads the same unmodified query image.
Relying entirely on the LVLM's internal perception, these methods add no visual signal to compensate for the perceptual failures that verification in this setting exposes.

\subsection{Visual Scaffold for LVLMs}
Visual scaffolds compensate for perceptual limitations of LVLMs by
adding training-free visual information onto images and have proven
effective across a range of vision
tasks~\cite{yang2023setofmark,hu2024visualsketchpad}.
However, existing scaffolds derive their parameters from image space and do not reflect the constraint structure that defines logical anomalies.
\emph{Disambiguation scaffolds} such as SoM~\cite{yang2023setofmark} make duplicate instances visually distinguishable, yet their ID numbering follows the spatial order of the image and is unrelated to the binding structure of
the constraint.
\emph{Alignment scaffolds} provide a cross-image reference frame for continuous measurements: SCAFFOLD~\cite{lei2025scaffold} overlays a coordinate grid, VISER~\cite{izadi2026visual} numbered horizontal lines, which we adapt as a single-axis reference, and Bbox+coords a box-and-coordinate form we adapt from the drawing tools of Visual Sketchpad~\cite{hu2024visualsketchpad}. In all three, the spacing and offset are fixed and independent of the normal range defined by few-shot references.
Visual Sketchpad itself selects tools \emph{agentically} from query-time perception, which in LAD is self-defeating: the scaffold is needed precisely because that perception is unreliable.

%% file: sections/03_preliminaries.tex
\begin{figure*}[!t]
\centering

\includegraphics[width=\textwidth]{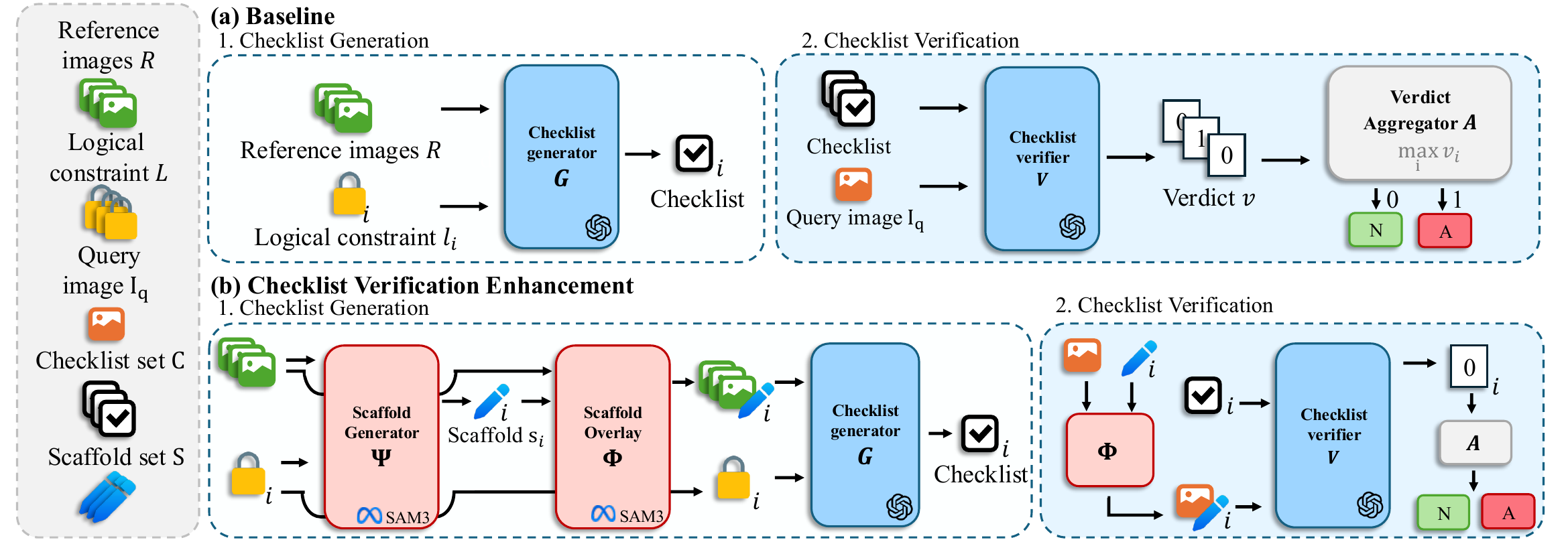}

\caption{Method comparison between the baseline and {\method}. Both methods use checklist generator and checklist verifier for verdicts. Unlike the baseline, ours overlays a constraint-specific scaffold on the query image, so each constraint is verified with an adapted visual input.}
\label{fig:framework}
\end{figure*}

\section{Preliminaries}
\label{sec:prelim}

Logical anomalies are defined as violations of category-specific
constraints~\cite{bergmann2022beyond}. A reference image is a normal sample
that defines what a correct instance of the category looks like. We write
$R = \{I_{r_1}, \dots, I_{r_K}\}$ for the set of $K$ reference images, and we
set $K{=}3$ throughout our experiments. A query image $I_q$ is the test-time
input whose normality must be decided. Given $R$ and $I_q$, LAD decides whether
$I_q$ violates any of the category's constraints.

Each category is governed by a set of logical constraints
$L = \{l_1, \dots, l_M\}$, where $M$ is the total number of constraints, and
each constraint $l_i$ states a normal condition on one property of the category.
We group them into two kinds according to the operation their verification
demands. A \emph{binding} constraint concerns discrete structure and is checked
by counting or matching, such as requiring exactly one pushpin per compartment.
A \emph{calibration} constraint concerns a continuous measurement checked by
comparison against the references, such as a cable length matching the reference
instances. This binding--calibration split organizes our analysis in
Section~\ref{sec:analysis}.

% Logical anomalies are defined as violations of category-specific
% constraints~\cite{bergmann2022beyond}. Attributing an anomaly to the constraint
% it violates requires those constraints to be explicit inputs to the detection
% system. A reference image is a normal sample that defines what a correct
% instance of the category looks like. We write $R = \{I_{r_1},\, \ldots,\,
% I_{r_K}\}$ for the set of $K$ reference images, and we set $K = 3$ throughout
% our experiments. A query image $I_q$ is the test-time input whose normality
% must be decided. Given $R$ and $I_q$, LAD decides whether $I_q$ violates any of
% the category's constraints.

% Not all constraints are verified in the same way, and this difference in the
% underlying perceptual operation motivates our approach. Each category is
% governed by a set of logical constraints $L = \{l_1,\, \ldots,\, l_M\}$, where
% $M$ is the total number of constraints, and each constraint $l_i$ states a
% normal condition on one property of the category. We group them into two kinds
% according to the operation their verification demands. A binding constraint
% concerns discrete structure and is checked by counting or matching, such as
% requiring exactly one pushpin per compartment or a correct pairing between
% groups. A calibration constraint concerns a continuous measurement and is
% checked by comparison against the references, such as a cable length matching
% the reference instances or a liquid fill level falling within the reference
% range. This binding--calibration split organizes our analysis in
% Section~\ref{sec:analysis}.

%% file: sections/04_method_new.tex
\section{Framework}
\label{sec:method}

Existing LVLM-based methods verify each constraint independently through a checklist-based framework~\cite{kwon2025logicqa, jin2025logicad,zhang2025logsad, zhang2024logicode}, organized in three stages: checklist generation prepares each constraint into a form an LVLM can evaluate from the constraint text and the references; checklist verification evaluates each prepared item against the query to produce a verdict; and verdict aggregation combines the per-constraint verdicts into an image-level decision, anomalous if even one constraint is violated.
 
% \subsection{Baseline: Three-phase Framework}

\subsection{Baseline: LVLM-based Framework}
\label{subsec:framework-baseline}

\paragraph*{Checklist Generation}
A generator $G$ converts each constraint $l_i$ into an independently 
evaluable item $c_i$ using the reference set $R$, yielding the 
checklist $C = \{c_1, \ldots, c_M\}$ via 
$c_i = G(l_i, R)$ for $i = 1, \ldots, M$. Existing methods differ only in the output format---a structured JSON
record of target object, property, and normal condition in
LogicAD~\cite{jin2025logicad}, a natural-language yes-or-no question in
LogicQA~\cite{kwon2025logicqa}---and in both, each item is converted
independently and is self-contained for verification.

\paragraph*{Checklist Verification}

A verification operator $V$ evaluates each item against the query image in a single LVLM call and returns a binary verdict $v_i$:
\begin{equation}
\label{eq:base-ver}
v_i = V(c_i,\, I_q), \qquad i = 1, \ldots, M.
\end{equation}
The text input $c_i$ varies per constraint, reflecting the per-constraint decomposition. The image input $I_q$, however, is the same raw query image for every constraint. Whether a constraint examines the count of identical objects or the length of a component, the call receives the identical, unmodified image. The question is tailored to the constraint, but the visual evidence is not.

\paragraph*{Verdict Aggregation}
Because a product is defective if it violates even a single constraint, the query image is classified as anomalous whenever at least one constraint is violated.
Given binary constraint-level verdicts $\{v_i\}_{i=1}^{M}$, where $v_i=1$ indicates a violated constraint, we define the image-level abnormal verdict as $A(I_q)=\max_{1 \leq i \leq M} v_i$, where $A(I_q)=1$ denotes an abnormal image and $A(I_q)=0$ denotes a normal image.
AUROC evaluation requires a continuous score, for which we use the violated fraction $\mathrm{Score}(I_q)=\frac{1}{M}\sum_{i=1}^{M} v_i$.

% \subsection{SOLVE}
\subsection{Checklist Verification Enhanced Framework}

\label{subsec:framework-solve}

The decomposition introduced by the checklist applies to the text alone.
Checklist verification enhancement extends it to the image: each verification
call reads a scaffolded query in place of the raw query, with no model
parameters updated.

This enhancement adds two operators to the baseline pipeline while leaving
generation and verification otherwise intact. A scaffold-generation operator
$\Psi$ reads a constraint and the references and returns a scaffold
configuration, and a scaffold-rendering operator $\Phi$ overlays that
configuration onto an image. 
Given $l_i$ and $R$, $\Psi$ produces the configuration $s_i=\Psi(l_i,\, R)$, and $\Phi$ renders it onto each reference as $\hat{I}^i_{r_k}=\Phi(I_{r_k},\, s_i)$, forming the scaffolded reference set $\hat{R}^i=\{\hat{I}^i_{r_1},\, \ldots,\, \hat{I}^i_{r_K}\}$.

% \begin{equation}
% \label{eq:solve-gen}
% s_i = \Psi(l_i, R), \qquad
% \hat{I}^i_{r_k} = \Phi(I_{r_k}, s_i), \qquad
% \hat{R}^i = \{\hat{I}^i_{r_1}, \ldots, \hat{I}^i_{r_K}\}
% \end{equation}
%
The checklist item is then generated from the constraint and the scaffolded references as $c_i = G(l_i,\, \hat{R}^i)$ for $i = 1, \ldots, M$, so that the question matches what the scaffold exposes.
This mirrors the baseline checklist generation, but replaces the raw reference set $R$ with the scaffolded reference set $\hat{R}^i$.
%
% \begin{equation}
% \label{eq:solve-genc}
% c_i = G(l_i, \hat{R}^i), \quad i = 1, \ldots, M
% \end{equation}
% %

At verification time, $\Phi$ overlays the same configuration onto the query, producing a scaffolded query $\hat{I}^i_q$ tailored to constraint $l_i$, which the verifier reads in place of the raw query:
\begin{equation}
\label{eq:scaffold-ver}
\hat{I}^i_q = \Phi(I_q,\, s_i), \qquad
v_i = V(c_i,\, \hat{I}^i_q), \qquad i = 1, \ldots, M.
\end{equation}
Whereas the baseline tailors only the text per constraint, this enhancement
tailors both the text and the image, so that the per-constraint decomposition
applies to the visual evidence as well. Verdict aggregation is unchanged from
the baseline.
% , with the per-constraint verdicts combined by Eq.~\eqref{eq:base-ver}. 
The scaffold form follows the constraint type: a disambiguation scaffold for a binding constraint and an alignment scaffold for a calibration constraint. Section~\ref{sec:analysis} establishes which form each type requires, and Sections~\ref{sec:solve} and~\ref{sec:solve++} develop how $\Psi$ configures it.
 

%% file: sections/05_analysis_new.tex
\section{Reasoning and Perception Guides for the Verification Phase}

\label{sec:analysis}

In this section, we analyze the two guides available for the verification
phase\rev{:} a textual reasoning guide that reasons more carefully over the
checklist item\rev{,} and a visual guide that modifies the query image the verifier
reads. We empirically test the available forms of each guide across binding
and calibration constraints, establishing which overlay each constraint type
requires. These findings fix the scaffolds that \ours\ implements.

We adopt
GPT-5-image-mini~\cite{singh2025openai} as the single backbone and report
per-constraint Balanced Accuracy~\cite{brodersen2010balanced},
$\mathrm{BA}=(\mathrm{TPR}+\mathrm{TNR})/2$, since each constraint yields a
single binary verdict, for which AUROC is undefined; we therefore report BA
at the constraint level and AUROC only at the image level. The probes in
Section~\ref{subsec:ana-twofail} are synthetic images that vary a single factor
in isolation.
The comparisons below use a subset of MVTec-LOCO's $16$ logical
constraints, chosen to isolate each failure mode most sharply; the full-system
evaluation on all of them appears in Section~\ref{subsec:exp-main-loco}.

\subsection{Effectiveness of Textual Reasoning Guide}
\label{subsec:ana-textual}

We first test whether a textual reasoning guide can aid the checklist
verification. Since the checklist item $c_i$ is the text input to the verifier,
a natural means of improving verification is to have the model reason more
deliberately over $c_i$ before committing to a verdict. Such reasoning may be
strengthened along two complementary directions, reasoning more thoroughly
toward the verdict and revising it post hoc. We evaluate one guide for each,
Chain-of-Thought~\cite{wei2022chain}, which prompts the model to reason step by
step before answering, and Self-Refinement~\cite{madaan2023self}, which allows
it to revise its own verdict. Both replace the item $c_i$ with a
reasoning-guided item $c_i'$ while the query image $I_q$ stays fixed, so the verdict becomes $v_i = V(c'_i,\, I_q)$.

\begin{figure}[!t]
\centering
\includegraphics[width=1\linewidth]{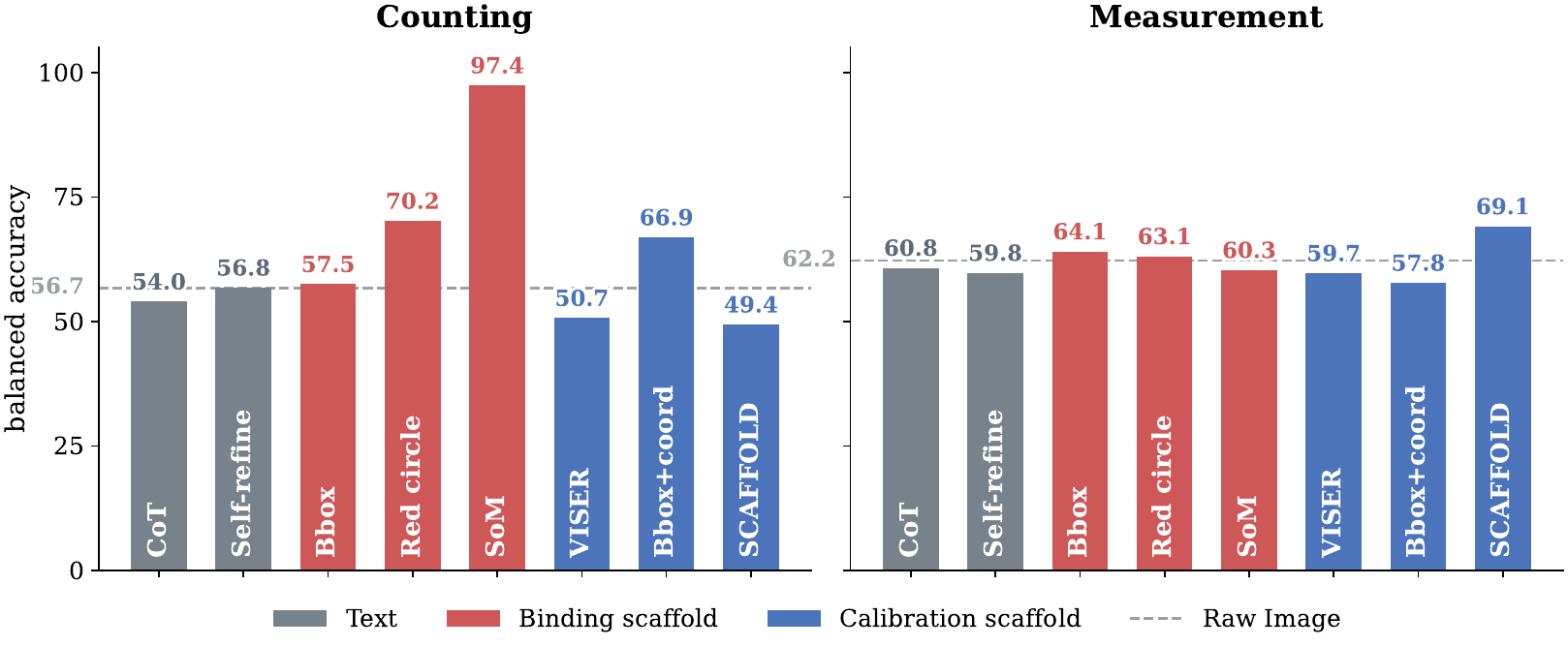} % count, measure 합침

\caption{Comparison of textual reasoning and visual perception guides. Bars show
balanced accuracy on a binding constraint for Counting and the average over
calibration constraints for Measurement, with the dashed lines indicating the
text-only baselines. Each overlay is applied to both constraint types, not
only the one it was designed for.}

\label{fig:placeholder}
\vspace{-8pt}
\end{figure}

Strengthening reasoning along either
direction, however, yields no measurable gain. As Fig.~\ref{fig:placeholder}
shows, neither guide improves on the text-only baseline by more than a
point on binding and calibration constraints alike. Reasoning more deliberately
over the same text therefore does not restore the verdict. If the limitation
does not lie in the reasoning, it must lie in what the model perceives from the
image, which we examine next.

\subsection{Effectiveness of Visual Guide}
\label{subsec:ana-twofail}

We next test whether a visual guide can recover what the textual guide could
not. A visual guide overlays a training-free cue onto the query image,
producing a guided query $\hat{I}^i_q$ while leaving the checklist item $c_i$
unchanged, so the verdict becomes $v_i = V(c_i,\, \hat{I}^i_q)$. This renders
the property a constraint examines directly observable to the verifier while
keeping the backbone and the question fixed. We evaluate the two overlay types
reported in prior work, a disambiguation overlay that assigns each instance a
separate visual identity and an alignment overlay that imposes a spatial
reference frame for continuous measurements. 

A single disambiguation overlay raises binding Count from $56.7$ to $97.4$
(Fig.~\ref{fig:placeholder}), a margin no textual guide approached. Since only the image changed while the
backbone and question remained fixed, the limitation is perceptual rather than
one of reasoning.

No single overlay, however, serves both: SoM reaches $97.4$ on Counting
but $60.3$ on Measurement, while SCAFFOLD reverses this ($49.4$ and $69.1$).
The two types therefore fail for distinct reasons, which we examine in turn.

\subsubsection{Instance Conflation on Binding Constraints}
\label{subsec:ana-duplication}

\begin{figure}[!t]
\centering
\includegraphics[width=\columnwidth]{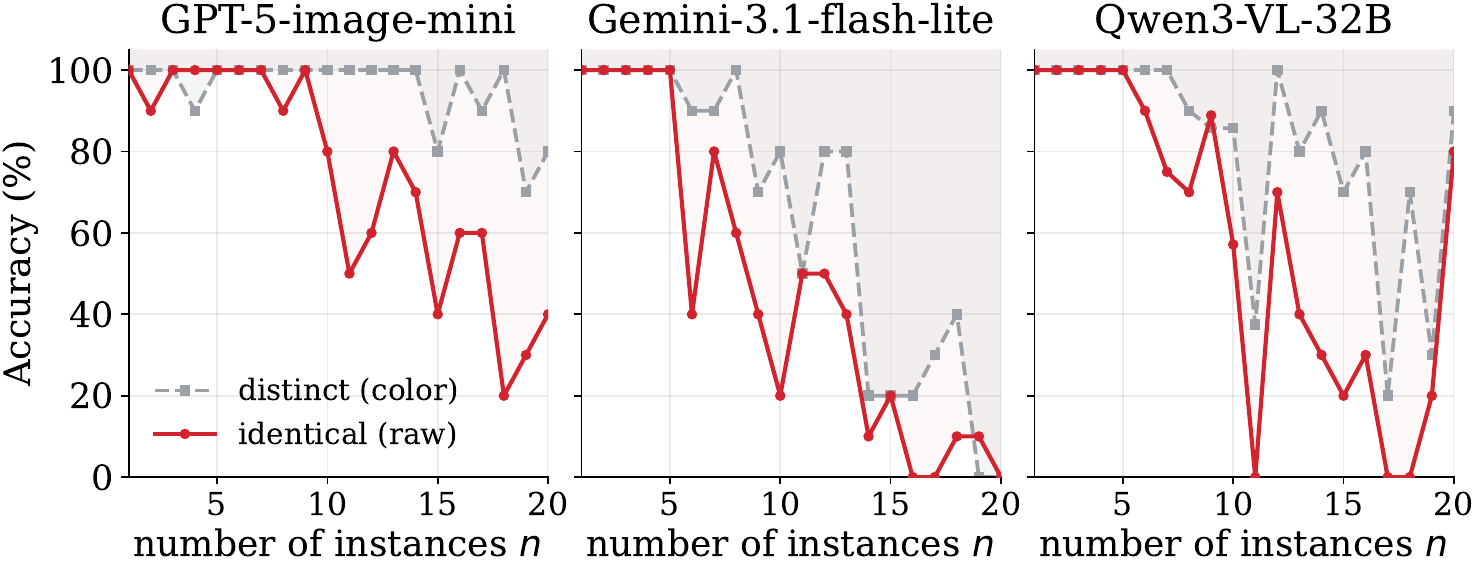}
\caption{Comparison of raw and per-instance colored images on a counting probe.
We query the count of $n$ identical circles on a blank canvas, varying only $n$.
Raw-image accuracy (red) collapses past $n \approx 10$, while distinctly
coloring each circle (gray) keeps it far higher throughout.}

\label{fig:counting}
\vspace{-8pt}
\end{figure}

We begin with the first perceptual guide type, a disambiguation guide for
binding constraints, where the verifier must individuate instances that look
alike. We ask whether providing a perceptual guide that makes each instance
visually distinct can restore this ability. Without such a guide, the model
registers shared attributes such as shape and color yet fails to bind them into
distinct instances~\cite{tong2024eyes, izadi2026visual}. If it could individuate
them, count accuracy would remain stable as their number grows. A synthetic
probe that counts $n$ identical circles shows the opposite: raw-image accuracy
collapses past $n \approx 10$ while per-circle coloring keeps it stable,
consistently across three backbones
(Fig.~\ref{fig:counting})~\cite{singh2025openai, gemini31flashlite,
bai2025qwen3}. Since color is the only factor altered, instance conflation is
structural rather than an artifact a larger backbone would resolve.

A per-instance color restores this separation, yet separation alone does not
determine the outcome. Among overlays that all separate the instances, accuracy
still diverges widely, from $70.2$ for a red circle to $97.4$ for SoM
(Fig.~\ref{fig:placeholder}). Separation is therefore necessary but
not sufficient, and what distinguishes these forms is examined in
Section~\ref{sec:solve++}. What the probe establishes is the overlay type: a
per-instance color is what a binding constraint requires, and \ours\ renders it
over SAM3 masks (Section~\ref{sec:solve}).

\begin{figure}[!t]
\centering
\includegraphics[width=\columnwidth]{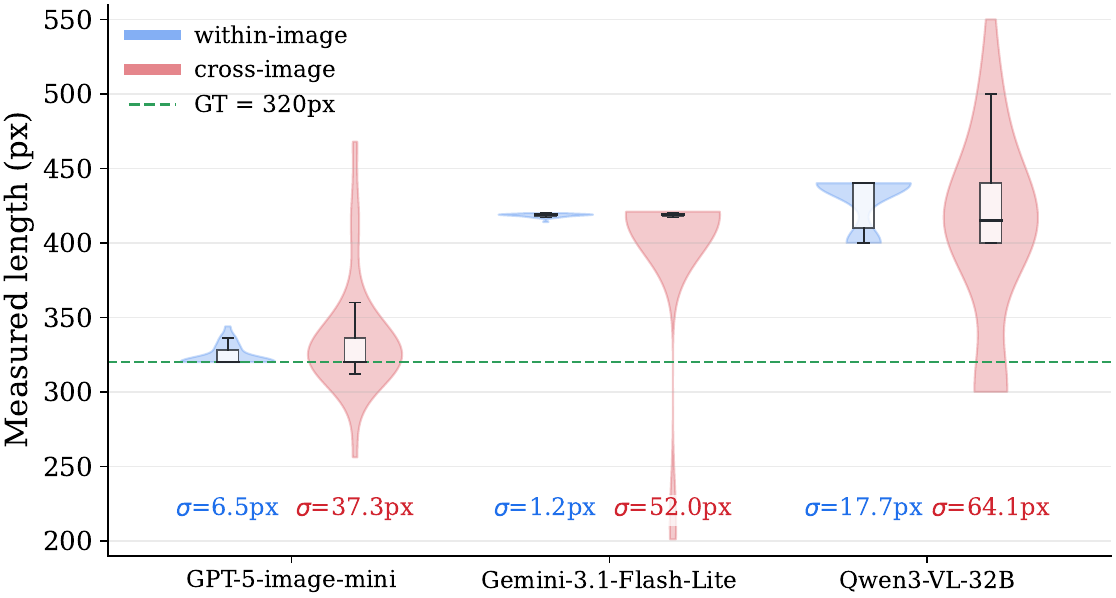}

\caption{Comparison of within-image and cross-image length readings on a scale
probe. We render a bar of fixed true length with a ground truth of $320$\,px and
read its length from the LVLM. A within-image reading repeats the same image,
while a cross-image reading spans separate renderings of the same length.}
\label{fig:scale-drift}
\vspace{-8pt}
\end{figure}

\begin{figure}[!t]
\centering
\includegraphics[width=\columnwidth]{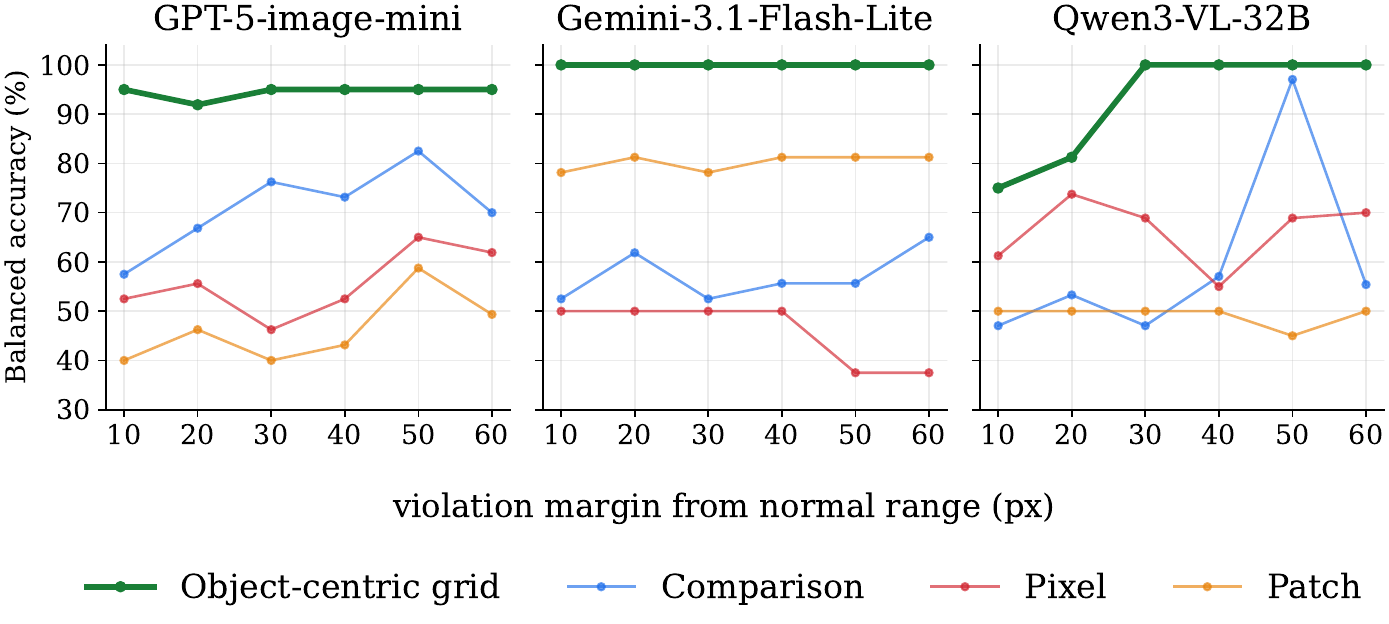}
\caption{Comparison of readout strategies on a range-membership probe. Balanced
Accuracy across violation margins for three VLM backbones. A test bar is judged
against the range of $K{=}3$ references, calibrated in code as in \ours, so the
backbone reads only the test: Comparison sees the references too and answers
directly, Pixel reports raw pixels, Patch units of an imagined $12{\times}12$
grid, and the object-centric readout an index off a scale drawn on the bar; the
last three are range-checked in code.}
\label{fig:measurement}
\vspace{-8pt}
\end{figure}

\subsubsection{Scale Drift on Calibration Constraints}
\label{subsec:ana-measurement}

We turn to the second perceptual guide type, an alignment guide for calibration
constraints, where the verifier must read the same physical length consistently
across the references and the query. The model struggles to do this because it holds no absolute scale and re-estimates the measurement on each image. If it held a consistent scale, the same length would read identically on whichever image carried it. Instead, reading a fixed-length bar across separate renderings spreads $3.6$ to $44$ times wider than reading it within one repeated image (Fig.~\ref{fig:scale-drift}), so the drift is structural rather than reading noise. The readings are also biased upward relative to the ground truth, but a constant bias is absorbed by a reference-relative grid, since the references and the query are read on the same scale; what such a grid cannot absorb is the cross-image spread. This drift is decisive near the normal-range boundary, where the object-centric grid sustains accuracy as the violation margin shrinks while the non-grid readouts drift (Fig.~\ref{fig:measurement}).

Removing the drift requires a grid that is discrete, two-dimensional, and object-anchored. A continuous coordinate would reinherit the drift, a single axis could not span a two-directional measurement, and a frame-fixed grid would shift cells as the object moves. The overlays that violate these requirements break down accordingly, as the single-axis VISER collapses on Length and the image-frame Bbox+coords collapses on Ratio, whereas the object-anchored grid holds across every calibration constraint. An object-anchored two-dimensional grid is thus the overlay a calibration constraint requires.

\medskip
The two probes identify the overlay each constraint type needs, a per-instance color for binding constraints and an object-anchored grid for calibration constraints. The overlay alone, however, does not settle the verdict. A color separates the instances but imposes no order on how they relate, and a grid frames the measurement but not at a resolution matched to any particular constraint. How each overlay must then be configured, from the constraint and from the references, is what \ourspp\ resolves (Section~\ref{sec:solve++}).

%% file: sections/06_SOLVE.tex
\section{CASCADE: Constraint-Aware Scaffolded Decomposition}
\label{sec:solve}

\ours\ implements the two overlays the analysis identified, routing each
constraint to the one its type requires. Where the checklist decomposes the
question per constraint, \ours\ decomposes the image per constraint, leaving all
model parameters unchanged.

\subsection{Pipeline}
\label{subsec:solve-pipeline}

\ours\ runs in four stages (Fig.~\ref{fig:framework}(b)). First, a parser reads the constraint $l_i$ and returns its type $\tau \in \{\text{binding}, \text{calibration}\}$ together with the target objects $T$ the constraint acts on. Second, $\Psi$ generates a type-matched scaffold configuration $s_i$, a disambiguation scaffold for a binding constraint and an alignment scaffold for a calibration constraint. Third, the checklist item $c_i$ is generated from the scaffolded references. Fourth, $\Phi$ overlays $s_i$ on the query, the item is verified in a single LVLM call, and a rule check returns the binary verdict $v_i$, aggregated as in the baseline.

Both scaffolds require grounding the target objects $T$ as masks\rev{.}
We adopt SAM3~\cite{ravi2025sam3}, a pretrained foundation segmentation model
used without any task-specific training, following the common practice of
rendering scaffolds on segmentation masks rather than raw
pixels~\cite{yang2023setofmark}\revb{.}

\subsection{Disambiguation Scaffold for Binding Constraints}
\label{subsec:solve-binding}

A binding constraint requires telling identical instances apart. The
disambiguation scaffold provides this by assigning a per-instance visual
identity. Each target set is segmented into one mask per instance
with SAM3, a distinct color is overlaid on each mask to delineate its extent,
and a numeric ID is written on top. The color separates the instances, and the ID
lets the verifier refer to them when a constraint relates rather than merely
counts them. In this stage $\Psi$ assigns the IDs sequentially in image order;
deriving them from the constraint is the subject of
Section~\ref{subsec:solvepp-numbering}.

\subsection{Alignment Scaffold for Calibration Constraints}
\label{subsec:solve-calibration}

A calibration constraint requires reading a continuous measurement on a scale that is consistent across images, which the analysis traced to scale drift and showed is removed by an object-anchored two-dimensional grid (Section~\ref{subsec:ana-measurement}). \ours\ recovers a rotated bounding box from the SAM3 mask, aligned to the object's principal axes, and draws the grid on this box so that the same physical endpoint maps to the same cell across the references and the query. The verdict is read as a comparison of cells: the references occupy a set of cells defining the normal range, and the query is judged normal when its endpoint falls in that range and a violation otherwise. In this stage $\Psi$ returns a single fixed cell width; calibrating it to the references is the subject of Section~\ref{subsec:solvepp-grid}.

%% file: sections/07_SOLVE++.tex
\section{CASCADE++}
\label{sec:solve++}

\ourspp\ replaces the two fixed defaults that $\Psi$ returns in \ours\ with
configurations derived from the constraint and the references. In each case it
is the derivation, not the overlay form, that determines the verdict.

\subsection{Constraint-Derived Numbering}
\label{subsec:solvepp-numbering}

The model does not merely read the IDs off the overlay but follows them as the
order in which to relate the instances, so the numbering must follow the
constraint's logical structure rather than the image. Color separates the
instances and raises \emph{Count} to $79.2$, but, lacking such an order,
recovers only part of \emph{Linking} ($61.4 \to 74.3$). A numeric ID supplies
one, yet where it leads is what decides the verdict: SoM's sequential ID runs
along image position rather than the constraint and leads the model toward the
wrong pairing, dropping \emph{Linking} to $66.7$, below color alone, while an
offset ID drops to $38.6$, below using no overlay at all
(Table~\ref{tab:disambig-config}).

$\Psi$ therefore derives the numbering from the constraint's logical structure
rather than the query image. Reading the constraint text and the parser's object
names, but not the query image, the LVLM returns only how the constraint
partitions the instances of $T$ into sets and under which correspondence between
them; the IDs then follow mechanically, all instances the correspondence
identifies with one another receiving a single index, and an instance it leaves
unmatched keeping its within-set ordinal. On Splicing Connectors the constraint
pairs the $n$ slots on one side with the $n$ on the other, so corresponding
slots carry the same ID and the verifier compares like-numbered slots across
sets, whereas SoM numbers the two sets $1,\dots,n$ and $n{+}1,\dots,2n$ and
leaves no such correspondence to follow. SoM and \ourspp\ differ only in what
the numbers denote, yet on \emph{Linking} this raises accuracy from $66.7$ to
$88.3$ (Table~\ref{tab:disambig-config}). What determines the verdict is not the
overlay's form but whether the correspondence it encodes is the one the
constraint asserts.

\begin{table}[!t]
\centering
\caption{Disambiguation channel decomposition for binding constraints on
Balanced Accuracy. We vary the two components of the overlay, color and a
numeric ID, together with the ID's numbering scheme, holding the question fixed
so that only the overlay changes. Count is the Pushpins (PP) count constraint,
and Linking the Splicing-Connector~(SC) slot-matching constraint.}

\label{tab:disambig-config}
\footnotesize
\setlength{\tabcolsep}{4pt}

\resizebox{\columnwidth}{!}{%
\begin{tabular}{llcc}
\toprule
Channel & Numbering & Count (PP) & Linking (SC) \\
\midrule
None              & ---                & $56.7$ & $61.4$ \\
Color only        & ---                & $79.2$ & $74.3$ \\
ID only           & offset             & $75.8$ & $38.6$ \\
ID only           & constraint-derived & $79.3$ & $86.2$ \\
SoM (color + ID)  & sequential         & $97.4$ & $66.7$ \\
Visual Sketchpad  & self-scaffolding   & $49.8$ & $69.5$ \\
\textbf{\ourspp\ (color + ID)} & \textbf{constraint-derived} & $\mathbf{99.6}$ & $\mathbf{88.3}$ \\
\bottomrule
\end{tabular}%
}
\end{table}

\subsection{Reference-Calibrated Grid}
\label{subsec:solvepp-grid}

An alignment grid must place its boundaries where the constraint's normal range
actually ends. Its cell width and position therefore have to follow the
reference distribution rather than a fixed default. The model judges a query
normal when its endpoint falls in the same cell as the references and a
violation otherwise, so the cell width sets the resolution of that judgment. A
cell narrower than the references scatter splits them across cells and flags a
normal query, while a cell wider lets a violation fall in the reference cell and
pass. The cell must therefore hold every reference reading and no more.

No fixed width meets this across constraints, because the references scatter by
different amounts depending on what the constraint measures. A two-way ANOVA
over cell widths and constraints (six widths $\times$ six constraints, five
reference sets each) confirms this, with the constraint-by-width
interaction accounting for the largest share of variance
($\eta^2 = 0.449$, $p < .001$). Two size-based rules---Dyn-size at half
the target's extent, Dyn-$2\times$ at a quarter---track the object rather than
the reference scatter, collapsing to $25.6$ on JB-Liquid and to chance on
SB-Length
(Table~\ref{tab:calib-rule}). The position matters equally, as a boundary
falling between the references splits the normal range even at the correct
width.

$\Psi$ therefore sets the grid from the reference distribution.
The analysis in Section~\ref{subsec:ana-measurement} showed that
LVLM measurements of the same physical quantity vary across images
by an amount that depends on the constraint. Let $r_k$ denote the
measurement of the target property from reference~$k$, and let
$\rho = \max_k r_k - \min_k r_k$ denote the spread across the $K$
references. \ourspp\ finds the narrowest cell of width $w$ and
offset $a$ that holds every $r_k$ with at least a margin~$m$ from
each boundary, judging a query normal when its reading lands in
this cell and a violation otherwise. Two backbone-dependent
constants set the resolution: the margin $m$, the smallest
separation the LVLM can reliably resolve between a reading and a
cell boundary, and the minimum width $w_{\min}$, the narrowest
cell at which it can still tell adjacent cells apart. Both are
calibrated once per backbone on synthetic probes and held fixed
across constraints ($m{=}25$\,px and $w_{\min}{=}60$\,px for
GPT-5-image-mini, the smallest values reaching $95\%$ probe
accuracy).

Concretely, the cell $[a,\,a{+}w]$ must satisfy
\begin{equation}\label{eq:calib-contain}
  \begin{aligned}
  & a + m \;\le\; r_k \;\le\; a + w - m,
    \quad \forall\, k \in \{1,\dots,K\},\\
  & w \;\ge\; w_{\min},
  \end{aligned}
\end{equation}
so that every reference reading sits at least a margin~$m$ from
either boundary. Eliminating~$a$, the tightest feasible width is
\begin{equation}\label{eq:calib-width}
  w^{\ast} \;=\; \max\bigl(w_{\min},\; \rho + 2m\bigr),
\end{equation}
and an offset that places every reference at least $m$ from either boundary is
\begin{equation}\label{eq:calib-offset}
  a^{\ast} \;=\; \min_k r_k \;-\; m.
\end{equation}
Together, $w^{\ast}$ and $a^{\ast}$ constitute the configuration $s_i$ that
$\Psi$ returns for a calibration constraint. Since $m$ and $w_{\min}$ are fixed
per backbone and only the references enter~$\rho$, no test-time reading touches
the calibration. When a constraint compares several groups or spans two
axes, the same containment conditions are solved jointly over a shared cell
width. The calibration is a
one-time cost per backbone: the constants reflect the backbone's measurement
noise, not any dataset or category, and are reused unchanged from MVTec-LOCO to
VID-AD (Section~\ref{subsec:exp-vidad}).
Not both constants need to be set precisely. The margin $m$ is a tolerance
the pipeline is largely insensitive to: varying it over a fourfold range around
the calibrated value moves balanced accuracy by at most $4.7$ points. The
minimum width $w_{\min}$ is the one that has to hold --- removing the floor lets
the cell shrink below what the backbone can read, and accuracy drops by more
than $35$ points.

\begin{table}[!t]
\centering
\caption{Calibration rule ablation on Balanced Accuracy (mean $\pm$ std over
five random reference selections). Grid parameterization per strategy: Fixed
$(w{=}100,\,a^{\ast})$, Dyn-size $(w{=}\text{tgt}/2,\,a^{\ast})$,
Dyn-$2\times$ $(w{=}\text{tgt}/4,\,a^{\ast})$, $a_{\mathrm{adv}}$
$(w^{\ast},\,\text{adversarial}~a)$, and \ourspp\ $(w^{\ast},\,a^{\ast})$.
Bold marks the best entry in each row.}
\label{tab:calib-rule}
\footnotesize
\setlength{\tabcolsep}{2pt}
\begin{tabular}{lccccc}
\toprule
Constraint & Fixed & Dyn-size & Dyn-$2\times$ & $a_{\mathrm{adv}}$ & \ourspp \\
\midrule
BB-Cereal & $77.6{\pm}9.9$  & $71.4{\pm}6.7$  & $78.6{\pm}10.1$ & $63.8{\pm}8.6$  & $\mathbf{82.9{\pm}4.0}$ \\
JB-Fruit  & $90.0{\pm}11.3$ & $\mathbf{95.9{\pm}1.9}$ & $76.9{\pm}6.1$ & $85.7{\pm}9.5$ & $94.4{\pm}3.9$ \\
JB-Text   & $83.3{\pm}6.6$  & $81.4{\pm}10.6$ & $82.6{\pm}2.3$  & $84.3{\pm}10.9$ & $\mathbf{96.9{\pm}0.1}$ \\
JB-Liquid & $\mathbf{97.8{\pm}1.0}$ & $25.6{\pm}7.9$ & $72.1{\pm}0.9$ & $84.0{\pm}4.3$ & $95.9{\pm}0.2$ \\
SB-Length & $92.2{\pm}0.4$  & $50.0{\pm}2.7$  & $\mathbf{96.4{\pm}0.5}$ & $86.7{\pm}2.0$ & $92.4{\pm}4.9$ \\
\midrule
\textbf{Avg.} & $88.2$ & $64.9$ & $81.3$ & $80.9$ & $\mathbf{92.5}$ \\
\bottomrule
\end{tabular}
\end{table}

\medskip
Both extensions follow one principle: the constraint type fixes the overlay,
while the constraint's logical structure and the reference distribution fix its
configuration.

%% file: sections/08_new_experiment.tex
\section{Experiments}
\label{sec:exp}

\subsection{Experimental Setup}
\label{subsec:exp-setup}

\begin{table}[!t]
\centering
\caption{Image-level AUROC (\%) on the MVTec-LOCO logical subset.
GPT-5 denotes GPT-5-image-mini and Gemini-3.1-FL denotes Gemini-3.1-Flash-Lite.
Rows without a marker are quoted from the
original papers; \textsuperscript{\dag} marks results we produced under each
method's own inference procedure, and \textsuperscript{*} marks a method whose design
admits only a single reference, where the other few-shot methods use $K{=}3$.
Bold marks the best entry in each column within each of the two regimes; dashes
mark entries the source paper does not report.}
\label{tab:main-loco}
\footnotesize
\setlength{\tabcolsep}{4pt}
\renewcommand{\arraystretch}{1.1}
\resizebox{\columnwidth}{!}{%
\begin{tabular}{ll cccccc}
\toprule
Method & Backbone & Avg & BB & JB & PP & SB & SC \\
\midrule
\multicolumn{8}{l}{\textit{Full-shot}}\\
PatchCore~\cite{roth2022patchcore}        & CNN & $74.0$ & $74.8$ & $93.9$ & $63.6$ & $57.8$ & $79.2$ \\
GCAD~\cite{bergmann2022beyond}            & CNN & $86.0$ & $\mathbf{87.0}$ & $\mathbf{100.0}$ & $97.5$ & $56.0$ & $89.7$ \\
EfficientAD~\cite{batzner2024efficientad} & CNN & $86.8$ & $85.5$ & $98.4$ & $\mathbf{97.7}$ & $56.7$ & $\mathbf{95.5}$ \\
AST~\cite{rudolph2023ast}                 & CNN & $79.7$ & $80.0$ & $91.6$ & $65.1$ & $\mathbf{80.1}$ & $81.8$ \\
LogSAD~\cite{zhang2025logsad}             & GPT-4o & $\mathbf{89.3}$ & --- & --- & --- & --- & --- \\
\midrule
\multicolumn{8}{l}{\textit{Few-shot}}\\
WinCLIP~\cite{jeong2023winclip}           & CLIP & $64.3$ & $57.6$ & $75.1$ & $54.9$ & $69.5$ & $64.5$ \\
LogicAD\textsuperscript{*}~\cite{jin2025logicad} & GPT-4o & $86.0$ & $93.1$ & $81.6$ & $98.1$ & $83.8$ & $73.4$ \\
                                          & GPT-5\textsuperscript{\dag} & $85.5$ & $94.7$ & $80.5$ & $97.6$ & $79.3$ & $75.3$ \\
LogicQA~\cite{kwon2025logicqa}            & GPT-4o & $87.6$ & $87.6$ & $88.2$ & $98.4$ & $71.5$ & $\mathbf{92.4}$ \\
                                          & GPT-5\textsuperscript{\dag} & $86.2$ & $91.6$ & $88.7$ & $97.3$ & $68.0$ & $85.3$ \\
LogSAD\textsuperscript{\dag}~\cite{zhang2025logsad} & GPT-4o & $84.6$ & $90.9$ & $85.4$ & $81.4$ & $81.7$ & $83.7$ \\
                                          & GPT-5\textsuperscript{\dag} & $86.6$ & $\mathbf{97.1}$ & $86.3$ & $84.1$ & $79.2$ & $86.6$ \\
\ours                                     & GPT-5 & $81.4$ & $90.0$ & $73.7$ & $93.8$ & $75.5$ & $73.8$ \\
\ourspp                                   & GPT-5 & $\mathbf{93.2}$ & $91.4$ & $\mathbf{94.7}$ & $96.6$ & $\mathbf{95.8}$ & $87.7$ \\
                                          & Gemini-3.1-FL & $92.7$ & $89.9$ & $92.9$ & $96.3$ & $95.2$ & $89.2$ \\
                                          & Qwen3-VL-32B & $90.6$ & $91.4$ & $84.5$ & $\mathbf{100.0}$ & $91.3$ & $85.6$ \\
                                          & Qwen3-VL-8B & $83.2$ & $87.7$ & $75.7$ & $82.1$ & $86.0$ & $84.4$ \\
\bottomrule
\end{tabular}%
}
\end{table}

We evaluate \ours\ on two benchmarks. MVTec-LOCO~\cite{bergmann2022beyond}
provides five product categories, Breakfast Box (BB), Juice Bottle (JB),
Pushpins (PP), Screw Bag (SB), and Splicing Connectors (SC), comprising $16$
constraints, $10$ binding and $6$ calibration; a single category may contain
both types. Per-constraint analyses name a constraint by its category and
measured target (e.g., JB-Liquid), and the measurement labels \emph{Count},
\emph{Ratio}, \emph{Length}, and \emph{Fill} denote PP-Count, BB-Cereal,
SB-Length, and JB-Liquid. VID-AD~\cite{nakata2026vidad} spans ten manufacturing
scenarios under five capture conditions (white, cable, and mesh backgrounds,
plus low-light and blurry), and its category specifications yield $20$
constraints, $16$ binding and $4$ calibration. All $36$ constraints across the
two benchmarks fall into one of the two types; the full list with its type
assignment is in our repository.

For MVTec-LOCO we adopt the natural-language constraints from
LogicQA~\cite{kwon2025logicqa} so that all vision-language methods share the
same constraint-input regime; for VID-AD we derive them from the dataset
authors' category specifications~\cite{nakata2026vidad}. The baselines were
originally reported on GPT-4o; to isolate scaffold design from backbone choice
we reproduce the publicly available ones, LogicAD~\cite{jin2025logicad} and
LogicQA~\cite{kwon2025logicqa}, on GPT-5-image-mini with their own inference
procedures and reference budgets intact. LogSAD~\cite{zhang2025logsad} routes
the task through specialized foundation models rather than a single swappable
chat LVLM, so we run the authors' released code under our $K{=}3$ protocol
instead of porting it.
The vision baselines are quoted from their original reports, as they are
trained on the full normal set and admit no few-shot variant.

\smallskip
\noindent\textbf{Implementation details.}
All LVLM calls use GPT-5-image-mini (\texttt{gpt-5-image-mini}) at high
reasoning effort, applied identically to our framework and the reproduced
baselines, with a single sample per call and no self-consistency. The parser runs once per category, not per query: we prompt the same backbone with the constraint text and the references, and it returns the constraint's
type, which decides the scaffold, together with the target objects and the text
prompt SAM3 needs to segment them. The type assignment matches our manual labels on all
$16$ MVTec-LOCO constraints, and this output is released rather than re-derived
at inference time. Object masks come from the official SAM3 checkpoint via
promptable concept segmentation on those prompts. All few-shot results draw
$K{=}3$ references uniformly at random with fixed seeds, and constraint-level
ablations (Tables~\ref{tab:disambig-config}--\ref{tab:calib-rule}) average over
five such selections against three for the image-level results
(Tables~\ref{tab:main-loco}, \ref{tab:main-vidad}, and \ref{tab:decomposition}).
Times are wall-clock on a single RTX A5000 at batch size one, excluding model loading. All prompts, code, and seeds are released in our repository.

% Cost

\begin{table}[!t]
\centering
\caption{Per-image inference cost (s) on MVTec-LOCO, all on GPT-5-image-mini.
\emph{Seg} covers each method's segmentation stage and \emph{LVLM} sums the
per-call API latencies; \emph{Total} is the full per-image wall clock, which
exceeds \emph{Seg}${+}$\emph{LVLM} by the remaining compute. LogSAD's checklist
runs once per category, so its per-image \emph{LVLM} is zero. The feature-based
detectors run locally in milliseconds and are not comparable on this axis.}
\label{tab:cost}
\footnotesize
\setlength{\tabcolsep}{16pt}
\renewcommand{\arraystretch}{1.15}
\begin{tabular}{lrrr}
\toprule
Method & Seg (s) & LVLM (s) & Total (s) \\
\midrule
LogicAD~\cite{jin2025logicad}  & $0.2$ & $52.6$  & $52.8$ \\
LogicQA~\cite{kwon2025logicqa} & $0.1$ & $186.5$ & $186.6$ \\
LogSAD~\cite{zhang2025logsad}  & $2.7$ & $0.0$   & $3.4$ \\
\ours                          & $3.2$ & $51.4$  & $55.0$ \\
\ourspp                        & $3.2$ & $68.3$  & $71.8$ \\
\bottomrule
\end{tabular}
\end{table}

\subsection{Main Results --- MVTec-LOCO}
\label{subsec:exp-main-loco}

% With three references \ourspp\ reaches $93.2{\pm}0.8$ AUROC over three
% reference sets without training \rev{(Table~\ref{tab:main-loco}). Under the
% same backbone and reference budget it improves over the strongest reproduced
% baseline (LogicQA\textsuperscript{\dag}, $86.2$) by $7.0$, and it exceeds the
% full-shot state of the art (LogSAD, $89.3$) by $3.9$ \revb{while using three
% references rather than the full normal set. Under our $K{=}3$ protocol the same
% method reaches $84.6$, so the margin over a shot-controlled comparison is
% $8.6$.}}

% With three references \ourspp\ reaches $93.2{\pm}0.8$ AUROC over three
% reference sets without training (Table~\ref{tab:main-loco}). Under the
% same backbone and reference budget it improves over the strongest reproduced
% baseline (LogSAD\textsuperscript{\dag}, $86.6$) by $6.6$, and it exceeds the
% full-shot state of the art (LogSAD, $89.3$) by $3.9$ while using three
% references rather than the full normal set.

With three references \ourspp\ reaches $93.2$ AUROC without training
(Table~\ref{tab:main-loco}); across three random reference sets the standard
deviation is $0.8$. Under the same backbone and reference budget it improves
over the strongest reproduced baseline (LogSAD\textsuperscript{\dag}, $86.6$) by
$6.6$, and it exceeds the full-shot state of the art (LogSAD, $89.3$) by $3.9$
while using three references rather than the full normal set.

Base \ours\ reaches only $81.4$, below both reproduced baselines. This is
the image-level counterpart of what Table~\ref{tab:disambig-config} shows at
the constraint level: a scaffold whose configuration misaligns with the
constraint misleads the verifier more than no scaffold at all. Deriving the
configuration from the constraint and the references adds $+11.8$ and reverses
the ordering. The gain concentrates on calibration-heavy categories, where the
reference-calibrated grid replaces the fixed default: Juice Bottle $+21.0$,
Screw Bag $+20.3$, and Splicing Connector $+13.9$.

% \revb{Image-level AUROC aggregates over sixteen constraints and conceals where
% each method actually fails. LogicAD and LogicQA differ by $0.7$ points in
% aggregate, yet diverge by over twenty on individual constraints in both
% directions. We therefore compare all sixteen in Balanced Accuracy against the
% same baselines (repository): the scaffold's gain concentrates on constraints
% where every text-only method falls below
% \todo{OPEN-1: fix the per-constraint threshold (60 or 75)}, and is absent where the
% raw image already resolves the constraint.}

% 09-09
% \revb{We also compared the methods at the constraint level, where the scaffold's
% effect separates into two regimes. LogicAD and LogicQA differ by $0.7$ points at
% the image level, yet diverge by over twenty on individual constraints in both
% directions, so the image-level figure hides which constraints each method fails.
% Against the same baselines in Balanced Accuracy, the scaffold's gain
% concentrates on constraints that no text-only method solves, and is absent where
% they already reach ceiling. The full comparison over all sixteen constraints is
% in our repository.}

LogicAD and LogicQA differ by $0.7$ points at the image level yet diverge
by over twenty on individual constraints in both directions, so the image-level
figure hides which constraints each method fails. Against the same baselines in
Balanced Accuracy, the scaffold's gain concentrates on constraints that no
text-only method solves and is absent where they already reach ceiling; the full
comparison is in our repository.

\ourspp\ is also the only framework that stays above $87$ on every
category, whereas every baseline collapses on at least one---PatchCore and
EfficientAD near $57$ on Screw Bag, LogicAD $73.4$ on Splicing Connector,
LogicQA $71.5$ on Screw Bag---since routing each constraint to its matched
scaffold keeps one failure mode from dominating the score.

The gain comes from the scaffold, not the backbone. Reproducing LogicAD and
LogicQA on GPT-5-image-mini yields $85.5$ and $86.2$, barely above their GPT-4o
results ($86.0$ and $87.6$): a stronger backbone reasons better but cannot
recover evidence the raw image never made visible. The same scaffold transfers
across backbones, holding $92.7$ on Gemini-3.1-Flash-Lite and $90.6$ on
Qwen3-VL-32B, and even the 8B model ($83.2$) approaches the full-shot vision
state of the art.

Table~\ref{tab:cost} places this gain against its cost. \ourspp\ verifies an
image in $71.8$\,s against LogicQA's $186.6$\,s on the same backbone, with
segmentation and rendering contributing $3.5$\,s of a total dominated by API
latency---well above the millisecond latencies of feature-based detectors, so
the framework targets near-line inspection rather than in-line control.

% \revb{The cost columns of Table~\ref{tab:main-loco} place this gain in context.
% \ourspp\ verifies an image in $71.8$\,s against LogicQA's $186.6$\,s on the
% same backbone, and the scaffold accounts for little of that: segmentation and
% rendering together contribute $3.54$\,s, under $5\%$ of a total dominated by API
% latency. That term shrinks with a lighter backbone but not to the millisecond
% latencies of feature-based detectors, so the framework targets near-line
% inspection rather than in-line control.}

\begin{table}[!t]
\centering
\caption{Main results on VID-AD (image-level AUROC, \%). W, C, M, L,
and B denote white, cable, mesh background, low-light, and blurry conditions,
respectively. Full-shot methods use the complete training set, whereas
\ourspp\ uses $K{=}3$ references with no training. Bold marks the best
entry in each column within each of the two regimes.}
\label{tab:main-vidad}
\footnotesize
\setlength{\tabcolsep}{3pt}
\renewcommand{\arraystretch}{1.1}
\resizebox{\columnwidth}{!}{%
\begin{tabular}{lcccccc}
\toprule
Method & W & C & M & L & B & Mean$\pm$Std \\
\midrule
\multicolumn{7}{l}{\textit{Full-shot}}\\
PatchCore~\cite{roth2022patchcore}        & $40.4$ & $45.0$ & $48.0$ & $46.9$ & $42.0$ & $44.5{\pm}2.9$ \\
EfficientAD~\cite{batzner2024efficientad} & $52.6$ & $60.2$ & $60.8$ & $47.9$ & $48.0$ & $53.9{\pm}5.7$ \\
PaDiM~\cite{defard2021padim}              & $34.8$ & $35.4$ & $42.8$ & $35.4$ & $40.3$ & $37.8{\pm}3.2$ \\
VAE~\cite{kingma2014vae}                  & $44.8$ & $49.4$ & $45.4$ & $47.2$ & $39.3$ & $45.2{\pm}3.4$ \\
AnoGAN~\cite{schlegl2017anogan}           & $48.9$ & $54.8$ & $49.2$ & $47.4$ & $46.8$ & $49.4{\pm}2.8$ \\
CSAD~\cite{hsieh2024csad}                 & $69.3$ & $62.4$ & $64.1$ & $69.4$ & $65.7$ & $66.2{\pm}2.8$ \\
VID-AD baseline~\cite{nakata2026vidad}    & $\mathbf{82.5}$ & $\mathbf{81.1}$ & $\mathbf{84.8}$ & $\mathbf{84.2}$ & $\mathbf{82.6}$ & $\mathbf{83.1{\pm}1.3}$ \\
\midrule
\multicolumn{7}{l}{\textit{Training-free}}\\
UniVAD~\cite{gu2025univad}                & $57.4$ & $59.2$ & $57.0$ & $57.3$ & $55.9$ & $57.4{\pm}1.1$ \\
\ourspp                                   & $\mathbf{94.7}$ & $\mathbf{91.4}$ & $\mathbf{97.1}$ & $\mathbf{93.0}$ & $\mathbf{96.2}$ & $\mathbf{94.5{\pm}2.1}$ \\
\bottomrule
\end{tabular}}
\end{table}

\subsection{Main Results --- VID-AD}
\label{subsec:exp-vidad}

We verify generalization by applying the identical pipeline, without any
parameter adjustment, to VID-AD (Table~\ref{tab:main-vidad}). \ourspp\ achieves
$94.5{\pm}2.1$, exceeding the strongest baseline ($83.1$) by $11.4$ points
while using only three references against the full training set the other
methods require. The gain is consistent across all five capture conditions,
including low-light ($93.0$) and blurry ($96.2$), where feature-based methods
degrade most: EfficientAD falls to $47.9$ and $48.0$ respectively. Because
the scaffold makes the constraint-relevant evidence explicit before the
verifier reads it, the capture condition affects the verdict far less than it
affects methods that must recover that evidence from raw features.

% \subsection{\revb{Robustness to Segmentation Quality}}
% \label{subsec:exp-segrobust}

% \revb{Both scaffolds are rendered on SAM3 masks, so a faulty mask could in
% principle corrupt any verdict downstream --- but it does so in one direction
% only. On MVTec-LOCO we inject boundary error ($\pm\delta$\,px dilation or
% erosion) and instance error (split, merge, drop) at the mask stage, leaving the
% references clean. The false-negative rate rises by at most $0.013$ across every
% condition, while false positives account for the whole of the accuracy loss.
% This follows from the decision rule: a corrupted mask moves the reading out of
% the reference range, and outside the range is a violation, so a normal image can
% be flagged but a violating one is not absolved. The errors are also not silent
% --- $94$ to $100\%$ of injected instance faults change the instance count.}

Table~\ref{tab:decomposition} independently enables or disables each scaffold.
The disambiguation scaffold raises the binding score from $72.6$ to $80.9$
($+8.3$) while calibration is essentially unchanged ($62.7 \to 59.9$).
% The disambiguation scaffold raises the binding score from $72.6$ to $80.9$
% ($+8.3$) \revb{while calibration does not benefit ($62.7 \to 59.9$)}.
Conversely, the alignment scaffold raises calibration from $62.7$ to $70.0$
($+7.3$) while binding remains unchanged ($72.6 \to 72.9$). This double
dissociation confirms that the two failure modes are independent and licenses
routing each constraint to a single matched scaffold. Combining both raises
overall AUROC from $74.5$ to $93.2$.

\begin{table}[!t]
\centering
\caption{Decomposition ablation on MVTec-LOCO (image-level AUROC, \%). All
rows use the \ourspp\ configuration rules and differ only in which scaffolds
are enabled; ``No scaffold'' verifies the same checklist on the raw query
image. The Binding and Calibration columns score each image by the violated
fraction of binding-only and calibration-only sub-constraints, and so miss
anomalies of the other type; Overall uses all sub-constraints, which is why it
exceeds either column alone.}
\label{tab:decomposition}
\footnotesize
\setlength{\tabcolsep}{4pt}
\resizebox{\columnwidth}{!}{%
\begin{tabular}{lccccc}
\toprule
Configuration & Disambig. & Alignment & Binding & Calibration & Overall \\
\midrule
No scaffold              & ---        & ---        & $72.6$ & $62.7$ & $74.5$ \\
Disambiguation only      & \checkmark & ---        & $80.9$ & $59.9$ & $81.8$ \\
Alignment only           & ---        & \checkmark & $72.9$ & $70.0$ & $79.5$ \\
\textbf{\ourspp\ (both)} & \checkmark & \checkmark & $\mathbf{83.8}$ & $\mathbf{74.5}$ & $\mathbf{93.2}$ \\
\bottomrule
\end{tabular}}
\end{table}
\subsection{Decomposition Ablation}
\label{subsec:exp-ablation}

\subsection{Reference Sensitivity}
\label{subsec:exp-refsens}

The calibration grid estimates its cell width $w^{\ast}$ from $K$
references, so $K$ governs how reliably the normal range is recovered. The
disambiguation scaffold derives its configuration from the constraint text and
is unaffected by $K$, so this analysis focuses on calibration constraints.

At $K{=}1$ a single reference cannot span the normal range, causing
$w^{\ast}$ to be underestimated. The resulting grid is overly narrow and
misclassifies normal queries as violations. JB-Fruit, for example, reaches only
$70.7$ balanced accuracy with a standard deviation of $22.5$ across reference
sets, and BB-Cereal only $71.8{\pm}12.6$. As $K$ increases to $3$, the
estimated $w^{\ast}$ converges toward its full-reference value and the
standard deviation contracts sharply. Accuracy follows the same trend, with
JB-Fruit rising to $94.4{\pm}3.9$ and BB-Cereal to $82.9{\pm}4.0$. The margin
$m$ in the calibration objective absorbs the residual from estimating the range
with finitely many references, which is why the parameter convergence
translates into stable accuracy. Increasing $K$ to $5$ or $10$ yields
negligible further gain, so we fix $K{=}3$, which balances
reference acquisition cost against detection accuracy.

\subsection{Sensitivity to Segmentation Quality}
\label{subsec:exp-segsens}

% 9월 10일
% Both scaffolds are rendered on SAM3 masks, which raises the question of how
% much of the gain depends on segmentation quality. We inject faults at the mask
% stage and re-run the full pipeline on Qwen3-VL-32B, whose clean-mask average is
% $90.6$, leaving the references untouched. Boundary error is largely absorbed:
% dilating every query mask leaves the average within noise at $10$ and $20$\,px
% and moves it only at $50$\,px, to $85.5$, with the loss confined to SB-Length,
% the constraint whose reading has the narrowest tolerance. This is what the
% discrete grid is for---a reading displaced by a few pixels still falls in the
% same cell, so the verdict does not change. Instance error is not absorbed in
% the same way. Splitting one instance in two drops the average to $74.8$ and
% merging the nearest pair to $79.2$, and the two fail differently: a split adds
% an instance the constraint did not license, so a normal query reads as a
% violation and only false alarms rise, whereas a merge removes the very instance
% a count or pairing rests on, so a violating query can read as normal and misses
% rise as well.

Both scaffolds are rendered on SAM3 masks, which raises the question of how much
of the gain depends on segmentation quality. We inject faults at the mask stage
and re-run the full pipeline on Qwen3-VL-32B, whose clean-mask average is $90.6$,
leaving the references untouched. Boundary error is absorbed: dilating every
query mask leaves the average within noise at 10 and 20\,px and moves it only at
50\,px, to $85.5$, with the loss confined to SB-Length, the constraint whose
reading has the narrowest tolerance. This is what the discrete grid is for---a
reading displaced by a few pixels still falls in the same cell, so the verdict
does not change. Instance error is not absorbed in the same way. Splitting one
instance in two drops the average to $74.8$ and merging the nearest pair to
$79.2$, and the two fail differently: a split adds an instance the constraint
did not license, so only false alarms rise, whereas a merge removes the instance
a count or pairing rests on, so misses rise as well.

%% file: sections/09_conclusion.tex
\section{Conclusion}
\label{sec:conclusion}

We show that the limitation of checklist-based LVLM verification for LAD is
perceptual rather than textual, manifesting as instance conflation and scale
drift. \ours\ extends checklist decomposition to the image through
constraint-specific scaffolds, and \ourspp\ derives each scaffold's
configuration from the constraint structure and the reference distribution,
reaching $93.2$ AUROC on MVTec-LOCO and $94.5$ on VID-AD without training.
Future work may extend the approach to constraints that combine both perceptual
operations.